%% file: main.tex
\documentclass[11pt]{article}

\usepackage[preprint]{acl}

\usepackage{times}
\usepackage{latexsym}
\usepackage[T1]{fontenc}
\usepackage[utf8]{inputenc}
\usepackage{microtype}
\usepackage{inconsolata}
\usepackage{graphicx}
\usepackage{amsmath,amssymb}
\usepackage{booktabs}
\usepackage{algorithm}
\usepackage{algpseudocode}

\newcommand{\method}{BRAPO}
\hypersetup{pdftitle={BRAPO: Alternating Belief Refinement and Program Optimization for Sample-Efficient LLM-Guided Program Search},pdfauthor={Chengzhang Yu, Yiming Zhang, Zhendong Chu, Zhanpeng Jin}}

\title{BRAPO: Alternating Belief Refinement and Program Optimization for Sample-Efficient LLM-Guided Program Search}

\author{Chengzhang Yu\textsuperscript{1,*} \quad Yiming Zhang\textsuperscript{2,*} \quad Zhendong Chu\textsuperscript{3,\ddag} \quad Zhanpeng Jin\textsuperscript{1,\dag} \\
\textsuperscript{1}South China University of Technology \quad \textsuperscript{2}University of Science and Technology of China \quad \textsuperscript{3}Squirrel AI \\
\texttt{zjin@scut.edu.cn} \quad \texttt{zc9uy@virginia.edu}}

\begin{document}
\maketitle
{\renewcommand{\thefootnote}{*}\footnotetext{Equal contribution.}}
{\renewcommand{\thefootnote}{\dag}\footnotetext{Corresponding author.}}
{\renewcommand{\thefootnote}{\ddag}\footnotetext{Co-corresponding author.}}

\input{sections/abstract}
\input{sections/introduction}
\input{sections/framework}
\input{sections/brapo}
\input{sections/experiments}
\input{sections/conclusion}
\input{sections/declaration}

\bibliography{references}

\end{document}

%% file: sections/abstract.tex
\begin{abstract}
Large language model (LLM)-guided program search learns mainly which programs perform well.
What it learns about why they perform well stays in text that no evaluation tests.
When evaluations are expensive, these untested beliefs limit sample efficiency, because they decide which candidates are evaluated.
We treat budget-constrained program search as two interdependent processes: refining beliefs about why programs perform well and optimizing programs.
We propose alternating belief refinement and program optimization, a framework in which each round refines predictive beliefs with inexpensive pilot experiments before the refined beliefs choose a candidate for formal evaluation.
\method{} implements this framework with LLM agents that state competing hypotheses, forecast the outcomes of their experiments in advance, and revise their hypotheses in light of the pilot outcomes.
On each of two combinatorial optimization tasks and two model-training tasks, \method{} reaches higher scores within 10 formal evaluations than ShinkaEvolve, AdaEvolve, and \method{} without belief refinement.
After the pilot experiments, its forecasts of the formal outcomes become more accurate, gaining 0.65 bits of logarithmic score per forecast on average.
\end{abstract}

%% file: sections/introduction.tex
\section{Introduction}
\label{sec:intro}

\begin{figure*}[t]
\centering
\includegraphics[width=\textwidth]{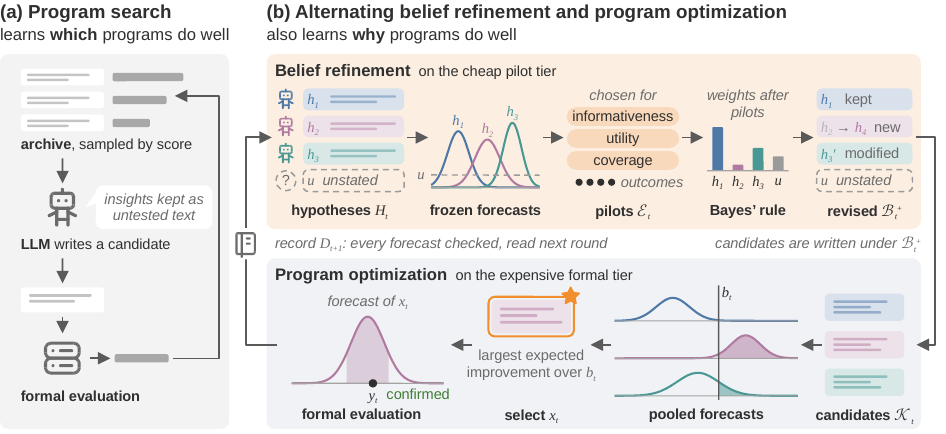}
\caption{Overview. (a) LLM-guided program search samples programs from an archive by score and keeps its insights as untested text. (b) Each round of alternating belief refinement and program optimization first refines the beliefs on the cheap pilot tier: the hypotheses and a reserve $u$ for unstated ones freeze their forecasts, pilot experiments test these forecasts, Bayes' rule reweights the hypotheses, and the hypotheses are revised into the refined belief state $\mathcal{B}_t^{+}$. Candidates are then written under $\mathcal{B}_t^{+}$, and the one with the largest expected improvement over the best score so far, $b_t$, goes to the expensive formal tier. The record $D_{t+1}$ carries every checked forecast into the next round.}
\label{fig:overview}
\end{figure*}

In program search guided by large language models (LLMs), a language model repeatedly proposes changes to a program, and each proposal is evaluated automatically \citep{romeraparedes2024mathematical,novikov2025alphaevolve,lange2026shinkaevolve,cemri2026adaevolve}.
Such a search learns mainly which programs perform well.
Its understanding of why they perform well is kept as text, such as summaries of earlier attempts or guidance for later generations \citep{lange2026shinkaevolve,cemri2026adaevolve}.
This text carries the search's beliefs back into the prompt, yet the beliefs make no prediction that an evaluation could check.
Untested beliefs suffice when evaluations are cheap, because the search can try many programs and keep the best.
FunSearch used on the order of a million samples \citep{romeraparedes2024mathematical}, and more recent systems still use on the order of a hundred evaluations per run \citep{lange2026shinkaevolve,cemri2026adaevolve}.

Recursive self-improvement \citep{schmidhuber2007godel,chen2026recursive} is turning program search toward the programs that train models, where each evaluation is itself a training run \citep{wang2026language,chen2026automated}.
The authors of one research agent judge trial and error over hundreds of attempts impractical for costly endeavors such as pre-training foundation models \citep{weng2026deepscientist}.
A search over such programs can evaluate far fewer candidates than it can write, so it must choose which to evaluate before seeing their scores.
Current searches choose mainly by the scores of programs already evaluated, building new candidates on high-scoring ones drawn from an archive \citep{romeraparedes2024mathematical,novikov2025alphaevolve,lange2026shinkaevolve,cemri2026adaevolve}.
A few observed scores say little about candidates not yet written, so each choice rests on untested beliefs about why programs perform well.
Systems that already work under such costs screen candidates with cheaper evaluations \citep{cheng2025language,li2026rldiscover} or predict their scores with learned models \citep{yu2026large,liu2026janus}.
Both kinds of signal estimate how well candidates will score and leave the beliefs untested.
We therefore ask whether LLM-guided program search can improve its sample efficiency by explicitly constructing, testing, and revising predictive hypotheses about why programs perform well.

We treat budget-constrained program search as two interdependent processes: refining beliefs about why programs perform well and optimizing programs.
A search framed this way looks for better programs and for better beliefs about how to find them.
We propose \emph{alternating belief refinement and program optimization} (Figure~\ref{fig:overview}).
Each round first refines the beliefs with inexpensive pilot experiments.
The search then writes candidates under the refined beliefs and sends the one with the largest expected improvement to an expensive formal evaluation.
The beliefs consist of hypotheses about where the best program found so far is limited.
Each hypothesis forecasts both the pilot scores of the ideas proposed in the current round and the formal scores of candidates written later, so pilot outcomes carry over to programs that do not yet exist.
Because language models typically overstate their confidence \citep{xiong2024can}, every forecast is frozen before its experiment and judged by its outcome.
Reweighting the hypotheses by Bayes' rule only shifts probability among those already stated \citep{murphy2026model}, so the search also revises the hypotheses themselves.
Language models make strongly correlated errors even across different models \citep{kim2025correlated}, so all stated hypotheses can miss the same explanation.
A reserve in the beliefs stands for the hypotheses that nobody has stated.

\method{} implements this framework with LLM researcher agents, which state competing hypotheses without seeing one another's.
To decide which ideas to pilot, it applies three principles: informativeness, utility, and coverage.
Across rounds, a record keeps every forecast, confirmed or contradicted, for the researchers to read.

We evaluate \method{} on two combinatorial optimization tasks, a route-planning problem from ALE-Bench \citep{imajuku2025alebench} and maximum satisfiability (MaxSAT) \citep{berg2024maxsat}, and on two model-training tasks, training a HalfCheetah policy in MuJoCo \citep{todorov2012mujoco,towers2025gymnasium} and a language model on enwik8 \citep{mahoney2006large}.
With the same 10 formal evaluations, \method{} reaches higher scores on all four tasks than ShinkaEvolve \citep{lange2026shinkaevolve}, AdaEvolve \citep{cemri2026adaevolve}, and \method{} without belief refinement.
On three of the four tasks, it exceeds the 10-evaluation scores of ShinkaEvolve and AdaEvolve within three formal evaluations.
The forecasts that decide the formal evaluations become more accurate after the pilots, gaining 0.65 bits of logarithmic score per forecast on average.
Our contributions are (i) the alternating framework, which casts budget-constrained program search as interdependent belief refinement and program optimization; (ii) \method{}, which implements the framework with LLM agents; and (iii) experiments on four tasks showing higher scores under the same formal budget and more accurate forecasts after the pilots.

%% file: sections/framework.tex
\section{Alternating Belief Refinement and Program Optimization}
\label{sec:framework}

\subsection{Problem Formulation}
\label{sec:problem}

Let $\mathcal{X}$ be the space of programs for a task.
The formal tier of a task is a fixed set of instances or seeds, each with a fixed resource limit.
Running $x\in\mathcal{X}$ on the formal tier returns its formal score $s(x)\in\mathbb{R}$, where higher is better.
We call one such run a formal evaluation.
The search starts from an initial program $x_0$ and may spend $T$ formal evaluations.
It chooses $x_1,\dots,x_{T-1}$ one at a time, and each choice can use only the scores observed before it.
Following a recent formulation of LLM-guided program search as score maximization under a computational budget \citep{cemri2026adaevolve}, we judge the search by the best formal score among the programs that it evaluates,
\begin{equation}
J=\max_{0\le t<T}s(x_t).
\label{eq:objective}
\end{equation}
We write $b_t=\max_{t'<t}s(x_{t'})$ for the best score before $x_t$ is chosen and call the program that attains it the current program $\hat{x}_t$.

A pilot experiment runs a program $x$ on the pilot tier and returns its pilot score $\tilde{s}(x)$.
The pilot tier is a cheaper version of the formal tier, with fewer instances or seeds, smaller resource limits, or a smaller model.
Pilot experiments do not count toward $T$.
The pilot experiments $\mathcal{E}_t$ between the formal evaluations of $x_{t-1}$ and $x_t$ have a total cost $c(\mathcal{E}_t)$ that must not exceed a pilot budget $C$.
Pilot scores do not enter $J$, so a pilot experiment can raise $J$ only by changing which programs are later evaluated formally.
As in multi-fidelity optimization \citep{kandasamy2019multi,klein2017fast}, we assume that pilot scores carry information about formal scores.

Formal evaluations are expensive, so $T$ is far smaller than the number of candidates that the search can write.
Each $x_t$ must be chosen by a prediction of $s(x_t)$.
Turning the scores of other programs into such a prediction requires beliefs about why programs perform well.

\subsection{Beliefs as Predictive Hypotheses}
\label{sec:beliefs}

In round $t$, which ends with the formal evaluation of $x_t$, the search starts from a belief state $\mathcal{B}_t=(H_t,q_t,D_t)$.
The belief state consists of a set $H_t$ of hypotheses, their weights $q_t$, and a record $D_t$ of earlier experiments, forecasts, and outcomes.
A hypothesis $h$ explains where the current program $\hat{x}_t$ is limited and names a measurement that would refute it.
An idea $a$ suggested by a hypothesis is either an improvement idea, which tries to raise the score of $\hat{x}_t$, or a diagnostic, which tests a hypothesis without trying to raise the score.

Each hypothesis forecasts two outcomes for every idea.
The first is the pilot score $Y^{\mathrm{P}}_a$, the value of $\tilde{s}$ reached by a program that implements $a$, with forecast $p(Y^{\mathrm{P}}_a\mid h)$.
The second is the formal score $Y^{\mathrm{F}}_a$ of a candidate built on $a$, with forecast $p(Y^{\mathrm{F}}_a\mid h)$.
Through $Y^{\mathrm{F}}_a$, a hypothesis predicts the formal scores of candidates before they are written.
A pilot experiment run now bears on programs that are written only later, through the hypotheses it favors.
This distinguishes belief refinement from screening, which uses cheap evaluations to rank candidates that already exist.

The search cannot state every explanation in advance.
The weights $q_t$ therefore also extend to a reserve $u$ for the hypotheses that have not yet been stated, with $q_t(u)=\lambda\in(0,1)$.
We call each $j\in H_t\cup\{u\}$ a component.
The belief state predicts each outcome $Y$, either $Y^{\mathrm{P}}_a$ or $Y^{\mathrm{F}}_a$, by the mixture
\begin{equation}
p(Y\mid\mathcal{B}_t)=\sum_{j\in H_t\cup\{u\}}q_t(j)\,p(Y\mid j),
\label{eq:mixture}
\end{equation}
where the weights of the stated hypotheses sum to $1-\lambda$.
Outcomes are independent given the component that generates them.
The reserve covers a range wider than all stated forecasts, so outcomes that every stated hypothesis rules out can still receive probability.
All forecasts are frozen before the experiments they concern are run.
Each hypothesis is therefore judged by the outcomes it predicted rather than by the confidence it reports.

\subsection{Alternating Rounds}
\label{sec:rounds}

\begin{algorithm}[t]
\caption{Alternating belief refinement and program optimization}
\label{alg:brapo}
\small
\begin{algorithmic}[1]
\Require $x_0$, $T$, $C$
\State Evaluate $x_0$ formally and start the record $D_1$
\For{$t=1,\dots,T-1$}
\Statex \hspace{\algorithmicindent}\textit{Belief refinement}
\State State $H_t$, $q_t$, and ideas; freeze $p(Y^{\mathrm{P}}_a\mid h)$ and $p(Y^{\mathrm{F}}_a\mid h)$
\State Choose $\mathcal{E}_t$ within $C$ by the three principles
\State Run $\mathcal{E}_t$ and reweight by Eq.~(\ref{eq:bayes})
\State Revise the hypotheses into $\mathcal{B}_t^{+}$
\Statex \hspace{\algorithmicindent}\textit{Program optimization}
\State Write $\mathcal{K}_t$; freeze $p(Y_x\mid\mathcal{B}_t^{+})$ for $x\in\mathcal{K}_t$
\State Select $x_t$ by Eq.~(\ref{eq:select}) and evaluate it formally
\State Append the forecasts and outcomes to form $D_{t+1}$
\EndFor
\State \Return the best program evaluated formally
\end{algorithmic}
\end{algorithm}

Rounds run for $t=1,\dots,T-1$, and each has two steps (Algorithm~\ref{alg:brapo}).

\paragraph{Belief refinement.}
In this step, the search keeps $\hat{x}_t$ fixed and chooses the pilot experiments $\mathcal{E}_t$ within the budget $C$ by three principles.
Informativeness favors ideas with a large mutual information $I(Z;Y^{\mathrm{P}}_a)$ under Eq.~(\ref{eq:mixture}), where $Z\sim q_t$ is the component that generates the outcome.
Bayes' rule moves weight only between components that forecast the outcome differently.
Because the components include the reserve, informativeness also favors ideas whose outcomes may fall outside every stated hypothesis.
Beliefs act on $J$ only through the candidates that are evaluated formally.
Utility therefore favors ideas with a large expected gain $\mathbb{E}[Y^{\mathrm{F}}_a\mid\mathcal{B}_t]-b_t$.
Candidates built on such ideas are likely to be evaluated.
Informativeness and utility are both computed from the forecasts in $\mathcal{B}_t$.
Coverage instead favors ideas that differ most from $\hat{x}_t$ and from the ideas already chosen, because outcomes of different approaches test the premises that the stated hypotheses share.
The three quantities are measured in bits, in score units, and as a dissimilarity, so each principle fills its own share of $\mathcal{E}_t$.

Let $\mathcal{A}_t$ be the ideas that $\mathcal{E}_t$ runs, and let $y^{\mathrm{P}}_a$ be the observed pilot score of $a\in\mathcal{A}_t$.
The search reweights the components by Bayes' rule,
\begin{equation}
w_t(j)\propto q_t(j)\prod_{a\in\mathcal{A}_t}p(y^{\mathrm{P}}_a\mid j),\qquad j\in H_t\cup\{u\}.
\label{eq:bayes}
\end{equation}
Weight that flows to the reserve flags outcomes that no stated hypothesis explains.
Reweighting cannot supply such an explanation, since it only redistributes weight within $H_t\cup\{u\}$.
The search therefore uses the outcomes to revise the hypotheses, keeping, modifying, or dropping each of them, or stating new ones.
The result is the refined belief state $\mathcal{B}_t^{+}$, which can contain hypotheses outside $H_t$.

\paragraph{Program optimization.}
Under $\mathcal{B}_t^{+}$, the search writes candidates $\mathcal{K}_t$ that change $\hat{x}_t$.
It forecasts the formal score $Y_x$ of each candidate $x\in\mathcal{K}_t$ by $p(Y_x\mid\mathcal{B}_t^{+})$, a mixture over the components of $\mathcal{B}_t^{+}$ in the form of Eq.~(\ref{eq:mixture}).
It then evaluates formally
\begin{equation}
x_t=\arg\max_{x\in\mathcal{K}_t}\mathbb{E}\big[(Y_x-b_t)_{+}\mid\mathcal{B}_t^{+}\big],
\label{eq:select}
\end{equation}
the candidate with the largest expected improvement over $b_t$ \citep{jones1998efficient}.
Let $y_t=s(x_t)$ be the formal score of $x_t$.
Since $b_{t+1}=\max(b_t,y_t)=b_t+(y_t-b_t)_{+}$ and $J=b_T$, Eq.~(\ref{eq:select}) chooses the candidate that the refined beliefs expect to raise $J$ the most in round $t$.
The forecasts and outcomes of the round are appended to the record to form $D_{t+1}$.
The hypotheses of round $t+1$ are based on $D_{t+1}$, so confirmed and contradicted forecasts accumulate across rounds.
Pilot experiments, which do not draw on $T$, are chosen for what they teach the beliefs.
Each formal evaluation is spent on raising $J$.

\subsection{Measuring Belief Refinement}
\label{sec:measure}

Frozen forecasts make belief refinement measurable.
Suppose that $x_t$ builds on an idea $a$ piloted in round $t$.
The formal score $y_t$ then has two frozen forecasts.
The pre-pilot forecast $p(Y^{\mathrm{F}}_a\mid\mathcal{B}_t)$ was made before the pilots and before $x_t$ was written.
The post-pilot forecast $p(Y_{x_t}\mid\mathcal{B}_t^{+})$ is the one that chose $x_t$.
With the logarithmic score \citep{gneiting2007strictly}, belief refinement in round $t$ is measured in bits by
\begin{equation}
\begin{split}
\delta_t={}&\log_2 p(Y_{x_t}=y_t\mid\mathcal{B}_t^{+})\\
&-\log_2 p(Y^{\mathrm{F}}_a=y_t\mid\mathcal{B}_t).
\end{split}
\label{eq:delta}
\end{equation}
Each probability in Eq.~(\ref{eq:delta}) is taken from a mixture that gives weight $0.99$ to the forecast and weight $0.01$ to a naive reference forecast built from the current score and the typical size of score changes on the same task.
The mixture keeps every logarithmic score finite.
A positive $\delta_t$ means that refinement made the beliefs that decide the formal evaluation predict its outcome better.

%% file: sections/brapo.tex
\section{\method{}}
\label{sec:brapo}

\begin{figure*}[t]
\centering
\includegraphics[width=\textwidth]{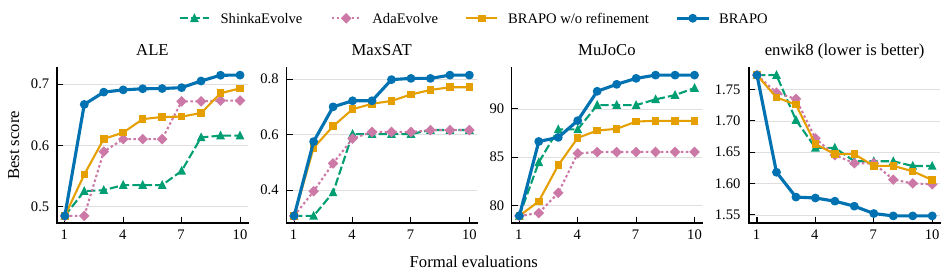}
\caption{Best score against the number of formal evaluations, averaged over three studies. On enwik8, we report bits per byte, where lower is better.}
\label{fig:curves}
\end{figure*}

\method{} implements the framework of \S\ref{sec:framework} with LLM agents and a host program that carry out the rounds of Algorithm~\ref{alg:brapo}.
Each researcher agent states one hypothesis, together with ideas and forecasts.
They submit at the same time and see no other submission until all are in, so no hypothesis is anchored on another's.
The host freezes every forecast before any code that it concerns runs.
Experimenter agents write the code for each selected idea and candidate.
The host runs this code within the resource limits of each tier and scores it.
It also applies every rule that selects an experiment or updates a weight.
The LLMs propose and interpret; explicit rules do all selection and scoring.

The host chooses the pilot experiments by the three principles of \S\ref{sec:rounds} and reweights the components by Eq.~(\ref{eq:bayes}).
It then publishes the pilot programs, their scores, and the updated weights to all researchers.
Each researcher revises its hypothesis in light of these results and proposes a candidate.
Every researcher then forecasts the formal score of every candidate.
The host pools these forecasts in the form of Eq.~(\ref{eq:mixture}), with equal weights on the researchers' forecasts and weight $\lambda$ on the reserve.
It selects $x_t$ from the pooled forecasts by Eq.~(\ref{eq:select}).
After the formal evaluation, the host compiles the record from the frozen forecasts and the measured outcomes.
Every researcher reads the record before stating the hypothesis of the next round.

%% file: sections/experiments.tex
\section{Experiments}
\label{sec:experiments}

\subsection{Setup}
\label{sec:setup}

\paragraph{Tasks.}
The combinatorial tasks are the route-planning problem of AtCoder Heuristic Contest 027 (AHC027) from ALE-Bench (hereafter ALE) and MaxSAT on instances from the MaxSAT Evaluation 2024.
The model-training tasks train a HalfCheetah policy in MuJoCo within a fixed number of environment steps and a byte-level language model on enwik8 within a fixed number of floating-point operations.
On ALE, the score is the mean ratio of a lower bound to the official score over the instances.
On MaxSAT, it is the mean score of the MaxSAT Evaluation 2024 over the instances.
On MuJoCo, it is the interquartile mean of the return over seeds \citep{agarwal2021deep}, normalized so that a random policy scores 0 and TQC \citep{kuznetsov2020controlling} trained for one million steps in RL Baselines3 Zoo \citep{raffin2020rl} scores 100.
On enwik8, it is the negative of the bits per byte.
Each task has its own formal tier and cheaper pilot tier (\S\ref{sec:problem}).

\paragraph{Baselines.}
The first baseline, \method{} w/o refinement, is \method{} without belief refinement.
It states no hypotheses and runs no pilot experiments, so it learns only from formal outcomes.
It writes, forecasts, and selects candidates as \method{} does.
The other two, ShinkaEvolve \citep{lange2026shinkaevolve} and AdaEvolve \citep{cemri2026adaevolve}, are evolutionary-search systems that choose what to evaluate mainly by score.
Each of them runs with its own code.
Every program that they evaluate, including the initial program, is run on the formal tier by the host of \method{} and counts as a formal evaluation.
All systems start from the same initial program.

\paragraph{Protocol.}
On each task, every system runs three independent searches, which we call studies.
Each study is compared on its first $T=10$ formal evaluations, including that of $x_0$.
On MuJoCo, a host fault cut three formal evaluations short, and the three affected studies are compared up to the formal evaluation before the fault.
Every agent of every system uses \mbox{GLM-5.3-Flash} \citep{zai2026glm53flash}.
Intervals are 95\% percentile intervals from 10{,}000 resamples of whole studies \citep{efron1986bootstrap}.

\subsection{Results}
\label{sec:results}

\input{tables/table_scores}

\method{} reaches the best score within 10 formal evaluations on all four tasks (Table~\ref{tab:scores}).
\method{} improves over \method{} w/o refinement on every task, for example from 0.771 to 0.815 on MaxSAT and from 88.8 to 93.5 on MuJoCo.
Figure~\ref{fig:curves} shows that the lead appears early.
On ALE, MaxSAT, and enwik8, \method{} exceeds the 10-evaluation scores of both evolutionary-search systems within three formal evaluations.

The gains come through the forecasts that choose the formal evaluations.
Pooled over the four tasks, 51 formal evaluations test candidates built on an idea piloted in the same round.
On their outcomes, the post-pilot forecast earns a higher logarithmic score than the pre-pilot forecast, by a mean $\delta_t$ (Eq.~(\ref{eq:delta})) of $+0.65$ bits per forecast, with a 95\% interval of $[+0.21,+1.06]$.

%% file: tables/table_scores.tex
\begin{table}[t]
\centering
\small
\setlength{\tabcolsep}{3pt}
\resizebox{\columnwidth}{!}{%
\begin{tabular}{@{}lcccc@{}}
\toprule
System & ALE $\uparrow$ & MaxSAT $\uparrow$ & MuJoCo $\uparrow$ & enwik8 $\downarrow$ \\
\midrule
ShinkaEvolve & 0.616{\scriptsize$\pm$0.108} & 0.617{\scriptsize$\pm$0.144} & 92.2{\scriptsize$\pm$1.7} & 1.628{\scriptsize$\pm$0.069} \\
AdaEvolve & 0.674{\scriptsize$\pm$0.037} & 0.617{\scriptsize$\pm$0.036} & 85.6{\scriptsize$\pm$6.6} & 1.599{\scriptsize$\pm$0.117} \\
\midrule
\method{} w/o refinement & 0.694{\scriptsize$\pm$0.062} & 0.771{\scriptsize$\pm$0.033} & 88.8{\scriptsize$\pm$2.2} & 1.606{\scriptsize$\pm$0.031} \\
\method{} & \textbf{0.715}{\scriptsize$\pm$0.067} & \textbf{0.815}{\scriptsize$\pm$0.064} & \textbf{93.5}{\scriptsize$\pm$4.9} & \textbf{1.548}{\scriptsize$\pm$0.009} \\
\bottomrule
\end{tabular}}
\caption{Best score within 10 formal evaluations, as the mean $\pm$ sample standard deviation over three studies. On enwik8, we report bits per byte, where lower is better.}
\label{tab:scores}
\end{table}

%% file: sections/conclusion.tex
\section{Conclusion}
\label{sec:conclusion}

Alternating belief refinement and program optimization moves LLM-guided program search from learning which programs perform well to learning why they do.
It refines predictive beliefs with pilot experiments before each formal evaluation.
\method{} shows that when formal evaluations are scarce, searching for better beliefs is an effective way to search for better programs.

%% file: sections/declaration.tex
\section*{Declaration of Generative AI and AI-Assisted Technologies in the Writing Process}

During the preparation of this work, the authors used AI to assist in drafting and refining the text and to help with part of the data verification.
After using this tool, the authors reviewed and edited all of the content and take full responsibility for the content of the publication.